\documentclass[pdflatex,sn-mathphys-num,iicol]{sn-jnl}

\usepackage{graphicx}%
\usepackage{multirow}%
\usepackage{amsmath,amssymb,amsfonts}%
\usepackage{amsthm}%
\usepackage{mathrsfs}%
\usepackage[title]{appendix}%
\usepackage{xcolor}%
\usepackage{textcomp}%
\usepackage{manyfoot}%
\usepackage{booktabs}%
\usepackage{algorithm}%
\usepackage{algorithmicx}%
\usepackage{algpseudocode}%
\usepackage{listings}%

\theoremstyle{thmstyleone}%
\theoremstyle{thmstyletwo}%

\theoremstyle{thmstylethree}%

\begin{document}

\title[Dynamic Routing MoE for Traffic Sign Recognition]{Hierarchically Decoupled Mixture-of-Experts for Robust Traffic Sign Recognition in Complex Driving Scenarios}



\author[1]{Mingxiao Wang}
\author[1]{Xiaozhen Qu}
\author[1,2]{Bolin Gao}
\author[2]{Tong Wang}
\author*[2]{Lei He}\email{helei2023@tsinghua.edu.cn}

\affil[1]{School of Automotive and Traffic Engineering, Liaoning University of Technology, Jinzhou, Liaoning 121001, China}
\affil[2]{State Key Laboratory of Intelligent Green Vehicles and Mobility, School of Vehicle and Mobility, Tsinghua University, Beijing 100084, China}


\abstract{Traffic sign detection is a fundamental component of environmental perception in autonomous driving and intelligent transportation systems. However, most existing detectors rely on static inference with globally shared parameters, limiting their ability to adapt to diverse and unstructured traffic scenarios. As a result, a single static model often struggles to simultaneously handle both clear near-range samples and challenging conditions such as distant small targets or adverse weather environments.
To address this limitation, we propose CBDES MoE TSR, a hierarchically decoupled heterogeneous mixture-of-experts(MoE) framework for traffic sign recognition. The proposed framework departs from the conventional globally shared parameter paradigm by introducing a heterogeneous You Only Look Once (YOLO) expert pool together with a lightweight gating network, enabling an image-level dynamic routing mechanism. Based on the semantic characteristics of the input image, the gating module selectively activates the most suitable expert model from the expert pool, enabling a shift from fixed parameter fitting to on-demand dynamic representation. This design enhances feature extraction capability for specific scenarios while maintaining controlled inference overhead.
Experimental results demonstrate that the proposed method achieves a remarkable balance between detection accuracy and efficiency on the composite traffic sign dataset. Specifically, our method attains an mAP50-95 of 76.8\%, yielding a 2.3\% improvement over the baseline method (74.5\%) while simultaneously reducing computational overhead by approximately 39.4\%. These findings robustly validate the effectiveness of the proposed approach.}




\maketitle

\section{Introduction}\label{sec1}

\begin{figure}[htbp]
\centering
\includegraphics[width=0.95\columnwidth]{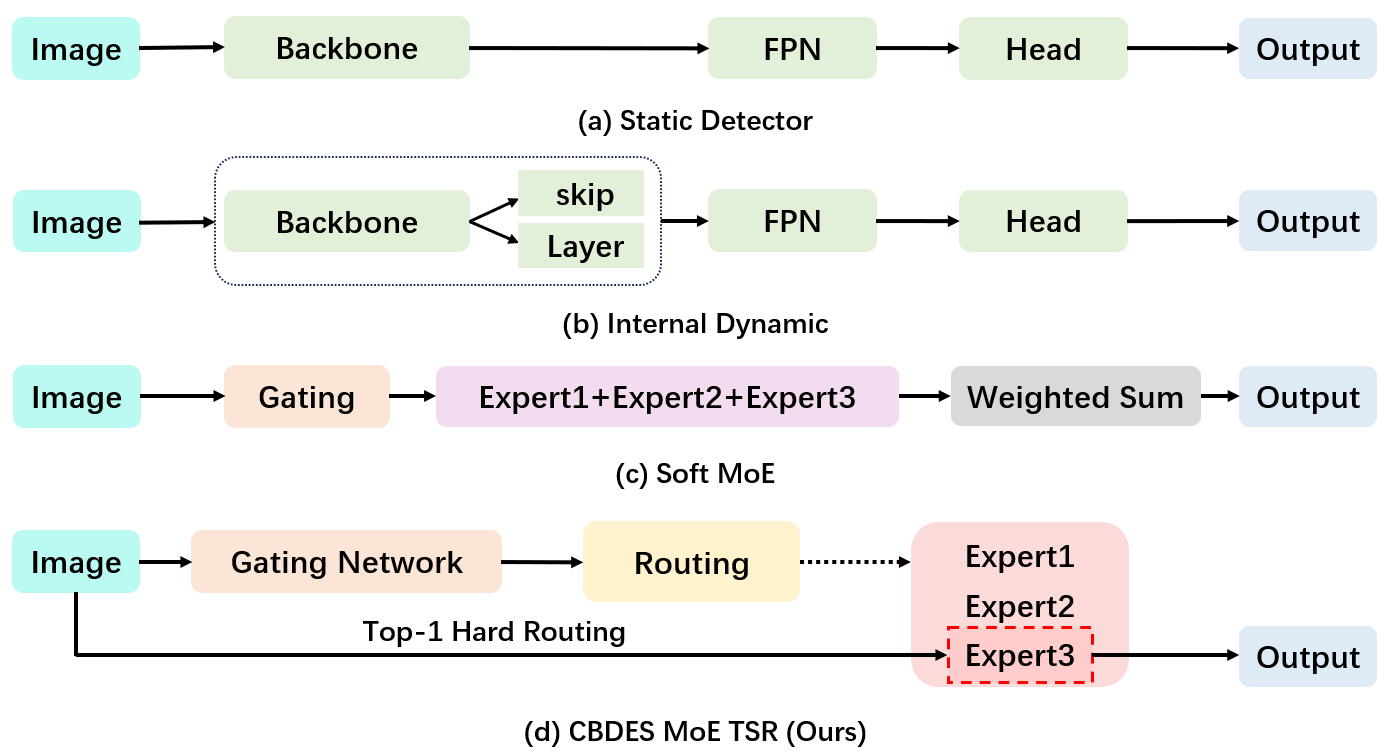} 
\caption{Comparison of different object detection paradigms. (a) Conventional static detector with a fixed computation graph. (b) Internal dynamic execution within the network through conditional layers. (c) Soft mixture-of-experts (MoE) where multiple experts are jointly activated and aggregated. (d) The proposed CBDES MoE TSR framework, which employs an image-level gating network to perform Top-1 hard routing over a heterogeneous expert pool.}\label{fig1}
\end{figure}

In recent years, environmental perception technologies for autonomous driving and intelligent transportation systems have achieved significant progress~\cite{ref1}. Classical static single-stage detectors integrate multi-scale feature extraction and deep hierarchical aggregation through highly optimized convolutional architectures and globally shared parameters~\cite{ref2, ref3}. As illustrated in Fig.~\ref{fig1}(a), these approaches have achieved competitive performance on standard public benchmarks through continuous improvements in backbone network structures and feature pyramid designs(FPN)~\cite{ref4}. However, in most cases, they lack adaptive perception capabilities tailored to the characteristics of input samples. Since all samples are processed using a fixed computational graph, these detectors often exhibit limited adaptability when facing unstructured and complex driving scenarios involving adverse weather disturbances or extreme scale variations~\cite{ref5, ref6}. Recently, mixture-of-experts models and dynamic neural networks have introduced conditional computation and sparse activation mechanisms~\cite{ref7, ref8}, providing a new perspective for building intelligent perception systems with both scenario adaptability and high representational efficiency.

Although mixture-of-experts models and dynamic neural networks have introduced conditional computation mechanisms that enable input-adaptive inference systems, effectively incorporating dynamic routing into real-time high-precision object detection remains challenging. One representative line of research, attempts to introduce dynamic pruning or conditional execution mechanisms within the network at the level of layers or channels in order to overcome the limitations of static computational graphs~\cite{ref9, ref10}, as illustrated in Fig.~\ref{fig1}(b). However, such fine-grained dynamic strategies often lead to irregular sparse structures, which may increase network design complexity and reduce training stability, particularly for object detection tasks.

Another line of work adopts a soft routing strategy~\cite{ref11} to enable expert-level conditional computation. As illustrated in Fig.~\ref{fig1}(c), multiple experts are activated simultaneously during inference and their outputs are combined through weighted aggregation. However, such fusion may weaken the structured inductive bias of individual experts, limiting functional specialization in object detection tasks.

To address this challenge, we propose a hierarchically decoupled mixture-of-experts framework for traffic sign detection, as shown in Fig.~\ref{fig1}(d), termed Component-Based Decoupled Expert System with Mixture-of-Experts for Traffic Sign Recognition (CBDES MoE TSR). The recent CBDES MoE \cite{ref12} successfully applied the hierarchical decoupling mechanism to bird's-eye-view (BEV) multi-modal perception tasks in autonomous driving. Unlike this method, which primarily serves cross-modal fusion and 3D object detection, our CBDES MoE TSR is specifically redesigned to tackle the extreme scale variations and adverse weather interference challenges faced by 2D traffic sign detection in unstructured scenarios. Inspired by the hierarchically decoupled paradigm of CBDES MoE \cite{ref12}, we argue that sample difficulty assessment and feature extraction in traffic sign detection should be functionally decoupled and processed in a hierarchical manner. Considering that different samples in real-world driving scenarios require varying depths of feature extraction, we introduce an expert pool composed of heterogeneous You Only Look Once (YOLO) detectors to construct a multi-dimensional feature representation space. Based on this design, an explicitly supervised image-level gating network is employed to guide input samples to the most appropriate expert model. This routing mechanism enables precise expert selection, which helps alleviate feature entanglement while establishing scenario-adaptive representation pathways, thereby improving feature extraction capability and detection stability in complex traffic environments.

Furthermore, efficient inference speed and reasonable computational cost are crucial for real-time perception in intelligent transportation systems, as these systems must respond promptly to rapidly changing road environments. Many existing high-accuracy approaches rely on computationally intensive heavyweight models to process all input scenarios. However, this static computation paradigm often leads to unnecessary computational overhead when handling a large number of simple near-range samples ~\cite{ref13}. To address this issue, we introduce a lightweight expert together with a lightweight gating network. By leveraging a sparse activation mechanism, the proposed design dynamically allocates low computational resources to frequently occurring simple samples, while high-capacity experts are activated only when necessary. In this way, computational cost can be further reduced while maintaining robustness across diverse scenarios, thereby improving the system’s average inference efficiency.

The main contributions of this work are summarized as follows:
\begin{itemize}
\item We propose CBDES MoE TSR, a hierarchically decoupled mixture-of-experts framework for traffic sign detection. It shifts from static parameter sharing to input-aware dynamic inference, overcoming the limited representational adaptability of conventional detectors in unstructured driving scenarios.
\item We introduce an explicitly supervised heterogeneous expert collaboration mechanism. By integrating diverse YOLO detectors, this strategy significantly improves detection robustness for challenging cases—particularly distant small objects—while reducing average computational costs.
\item Extensive experiments on a composite multi-domain dataset demonstrate the superiority of our approach. Compared with the YOLOv9c baseline, CBDES MoE TSR achieves a 2.3\% improvement in mAP50–95 and a 39.4\% reduction in computational cost, offering an optimal balance between accuracy and efficiency.
\end{itemize}

\section{Related Works}\label{sec2}

\subsection{Static Object Detection Frameworks for Traffic Sign Detection}\label{subsec2-1}

Traffic sign detection aims to simultaneously perform object localization and category recognition from road images, and serves as a fundamental component of perception modules in intelligent transportation systems and autonomous driving. In recent years, single-stage object detectors have been widely adopted for traffic sign detection due to their end-to-end training paradigm and favorable trade-off between detection accuracy and inference speed. Representative frameworks, such as the YOLO series, have continuously improved detection performance through advances in backbone architectures, feature fusion strategies, and training techniques. 
For example, YOLOv9~\cite{ref14} introduces programmable gradient information (PGI) to enhance feature learning, while YOLOv10~\cite{ref15}focuses on end-to-end optimization to improve deployment efficiency.

Despite their strong performance on standard benchmarks, these models fundamentally follow a static inference paradigm. During inference, the convolutional kernel weights and the computational graph topology remain strictly fixed. Consequently, regardless of whether the input corresponds to clear near-range traffic signs, extremely distant small-scale targets, or degraded weather conditions with reduced visibility, the network processes all samples using the same parameter distribution and computational path. While this unified parameter fitting strategy is simple and effective from an engineering perspective, it often fails to adapt to the significant variability of real-world driving scenarios. As a result, the dynamic representation capability of feature extractors across diverse conditions remains limited~\cite{ref6}. Therefore, overcoming the constraints of static computational graphs and introducing input-aware adaptive inference mechanisms is crucial for improving the environmental adaptability of detection systems.

\subsection{Dynamic Inference and Conditional Computation}\label{subsec2-2}

Dynamic inference and conditional computation introduce routing decisions during inference, allowing models to select different computational paths or subnetworks based on input features, thereby improving responsiveness to input variability under a limited computational budget~\cite{ref5}. Existing studies have explored conditional computation at different levels of granularity. One line of work performs fine-grained dynamic adjustments within the network, such as early exiting based on input difficulty or dynamic layer activation~\cite{ref16, ref17}. Another line of research introduces sparse routing at the model level, where a gating network assigns samples to a small subset of expert models, enabling the expansion of model capacity while keeping inference computation sparse and controllable. Representative works such as GShard and Switch Transformer~\cite{ref18, ref19} systematically investigate the training stability, parallelization strategies, and routing mechanisms of sparsely gated MoE models, providing a reproducible pathway for applying conditional computation in large-scale models.

However, in object detection tasks, fine-grained dynamic strategies often require intrusive modifications to the backbone network and feature pyramid structures, which may increase engineering complexity and introduce training instability. Unlike classification or language tasks, object detection relies heavily on multi-scale feature alignment and dense prediction architectures. As a result, dynamically changing computation paths may further complicate structural design and negatively affect training stability, potentially degrading localization accuracy. Therefore, introducing conditional computation while preserving the structural consistency of detection frameworks remains a unique challenge in object detection.

In comparison, model-level or module-level routing can be more easily integrated with existing detectors. By maintaining the structural integrity of individual detectors and the stability of their training paradigms, a lightweight gating network can determine which expert model should be activated for inference. This strategy enables on-demand computation with minimal architectural modifications, making it a practical approach for integrating conditional computation into object detection systems \cite{ref20, ref21}.

\subsection{MoE in Vision and Autonomous Driving Perception and the CBDES MoE Paradigm}\label{subsec2-3}

Mixture-of-Experts models have recently been widely explored in computer vision due to their ability to expand model capacity while maintaining sparse computation~\cite{ref22}. Several studies introduce MoE into vision tasks by adopting dense soft routing mechanisms, where the outputs of multiple experts are combined through weighted aggregation to integrate multi-domain features or enable cross-scenario adaptation~\cite{ref23}. In addition, some works adopt heuristic expert assignment strategies, for example routing inputs to different subnetworks according to target scale or image resolution. Such strategies can reduce implementation complexity while improving inference efficiency to some extent~\cite{ref24, ref25}.

In object detection, however, expert outputs must remain consistent not only in semantic representation but also in spatial localization. Consequently, compared with output fusion strategies, selecting a single expert for inference based on input semantics better satisfies the structural consistency requirements of detection tasks~\cite{ref20, ref26}. However, most existing methods still emphasize feature/prediction-level integration, leaving the coupling between routing and expert functionality weak and the functional boundaries unclear.
To address this, CBDES MoE introduces a hierarchically decoupled MoE framework at the functional-module level for autonomous driving perception, enabling sparse dynamic path selection with structurally heterogeneous experts while preserving each expert's structural integrity~\cite{ref12}. Experimental studies demonstrate that such a hierarchically decoupled design provides advantages when handling multi-modal and multi-complexity inputs, offering a new design perspective for applying MoE frameworks to autonomous driving perception tasks.

\section{Methodology}\label{sec3}

\subsection{Overall Architecture Overview}\label{subsec3-1}

\begin{figure*}[htbp]
\centering
\includegraphics[width=0.95\textwidth]{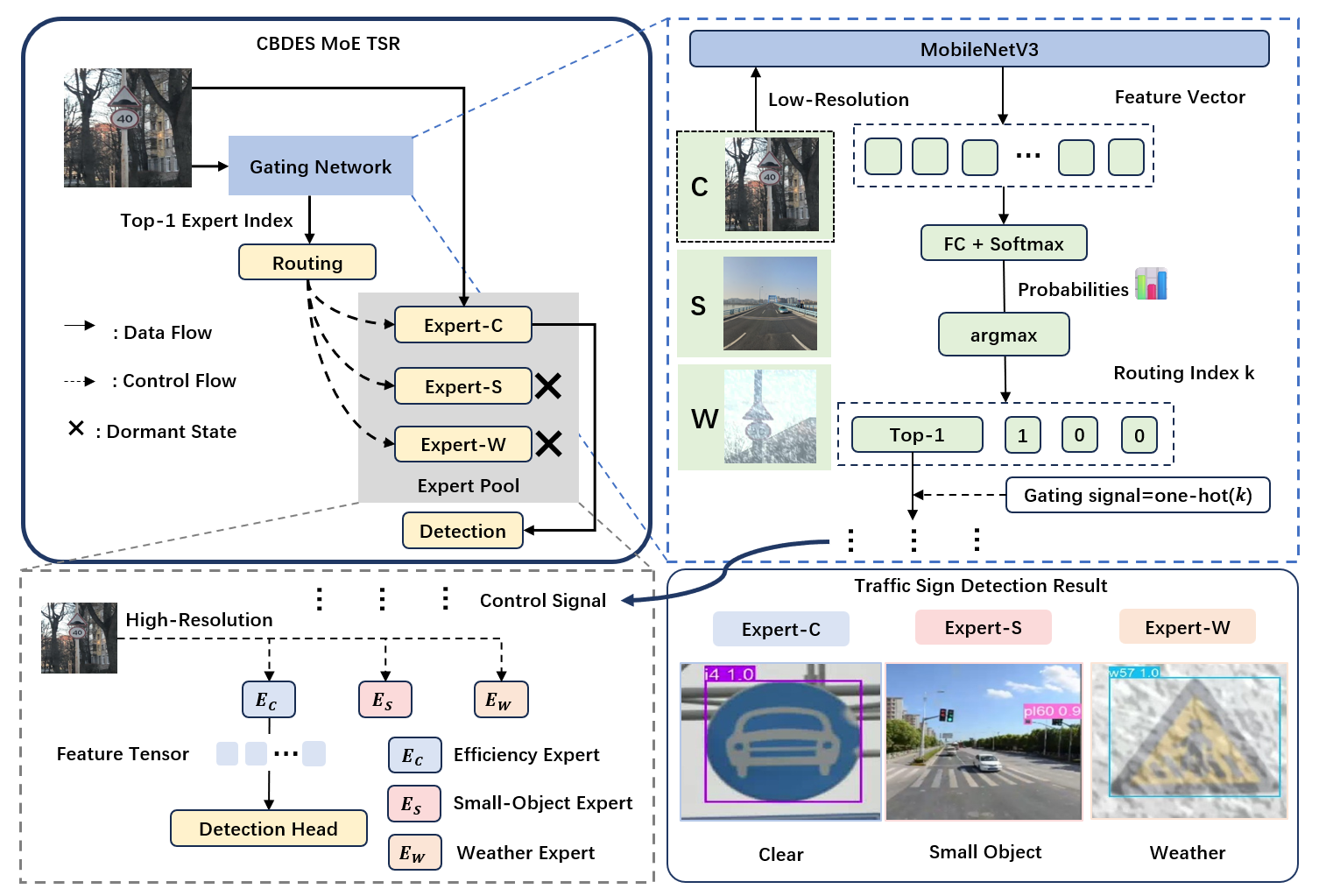}
\caption{Overall architecture of the proposed CBDES MoE TSR framework. The gating network analyzes a downsampled representation of the input image and produces a Top-1 routing index. Based on this index, a single expert from the heterogeneous expert pool (Expert-C, Expert-S, or Expert-W) is activated to perform detection on the original high-resolution image, while the remaining experts remain inactive. Solid arrows denote data flow, and dashed arrows indicate routing control signals.}\label{fig2}
\end{figure*}

In this paper, we propose CBDES MoE TSR, a heterogeneous mixture-of-experts detection framework based on a hierarchically decoupled design. The overall forward inference pipeline of CBDES MoE TSR is illustrated in the main block on the left side of Fig.~\ref{fig2}.

Specifically, given multi-scenario traffic images as input, CBDES MoE TSR first employs a lightweight semantic gating network, MobileNetV3-Small, to analyze the downsampled representation of the input image and identify its corresponding scene domain. Then, an explicitly supervised routing mechanism evaluates the semantic attributes of the input sample using a Top-1 hard routing strategy, producing a deterministic expert activation index.

The heterogeneous expert pool subsequently combines this activation index with the original high-resolution input image to perform sparsified conditional inference, activating only a single detector that best matches the current scene domain, such as standard scenes, small-object scenarios, or adverse weather conditions.

Finally, the selected expert model bridges the feature representation of the corresponding scenario with the object detection task, generating traffic sign detection results conditioned on a dynamically selected subset of model parameters. Through these components, CBDES MoE TSR effectively leverages dynamic routing to allocate expert modules on demand, thereby establishing an input-aware adaptive inference paradigm.

\subsection{Construction of the Heterogeneous Expert Pool}\label{subsec3-2}

To overcome the limitation of using a single fixed-parameter detector for all inputs, we construct a heterogeneous expert pool following the function-level decoupling principle of CBDES MoE (Fig.~\ref{fig2}). We decompose traffic-sign feature extraction into three specialized sub-directions and assign each to a dedicated expert with different capacity and architectural bias, namely $E_C$, $E_S$, and $E_W$, balancing representation diversity, routing stability, and model complexity.

\noindent\textbf{Efficiency Expert $E_C$}:
$E_C$ adopts a lightweight detector (YOLOv11s, Ultralytics~\cite{ref27}) with improved Cross Stage Partial with Kernel 2 (C3k2) modules and a compact feature pyramid, targeting clear short-range samples and maximizing inference throughput.

\noindent\textbf{Small-Object Expert $E_S$}:
$E_S$ employs a deeper YOLOv9c with Generalized Efficient Layer Aggregation Network (GELAN) and PGI to improve recall for distant small targets, compensating for the limitations of lightweight models in fine-grained feature extraction.

\noindent\textbf{Weather Expert $E_W$}:
$E_W$ also uses YOLOv9c but its core distinction lies in being trained on adverse-weather-augmented, all-scale data. It is engineered to provide robust feature representations tailored for complex and adverse weather scenarios such as rain, snow, and fog, thereby ensuring reliable perception performance for the system even in extreme environments.

Integrating these experts within a unified MoE framework yields a multi-bias representation space, enabling scenario-matched expert invocation and serving as the structural basis for the gating-based dynamic inference in the following section.

\subsection{Explicitly Supervised Gating Mechanism}\label{subsec3-3}
\label{subsec:gating}

To enable input-dependent dynamic expert selection, we design an explicitly supervised image-level gating mechanism. Unlike conventional MoE frameworks that learn routing strategies through implicit joint optimization, our approach directly supervises the gating network using predefined scene-domain labels, thereby improving the stability and controllability of the routing process. The input samples are partitioned into three mutually exclusive scene domains, and the gating label space is defined as follows:
\begin{equation}
    y \in \{C, S, W\}
\end{equation}
Here, $C$ denotes the standard clear scene domain, $S$ represents the small-object-dominant distant scene domain, and $W$ corresponds to the adverse-weather degradation domain.

The gating network adopts MobileNetV3-Small as a lightweight backbone to ensure that the computational overhead of routing decisions remains negligible. The input image is first downsampled to a low-resolution representation $I_{low}$. The gating network, parameterized by $\theta_G$, then extracts features from $I_{low}$ and outputs a scene-domain probability vector:
\begin{equation}
    p = \text{Softmax}(G(I_{low}; \theta_G)) \in \mathbb{R}^3
\end{equation}
where $G(\cdot)$ denotes the gating network, and the probability vector $p = [p_C, p_S, p_W]$ represents the predicted probabilities corresponding to the three scene domains $C$, $S$, and $W$.

During inference, a Top-1 hard routing strategy is adopted to determine the optimal expert assignment:
\begin{equation}
    k = \arg\max_{i \in \{C, S, W\}} p_i
\end{equation}

To formally describe the dynamic inference process of the CBDES MoE TSR framework, the final detection output $O(I)$ for a given high-resolution input image $I$ can be formulated as:
\begin{equation}
    O(I) = \sum_{i \in \{C, S, W\}} \mathbb{1}_{\{i = k\}} \cdot E_i(I; \theta_{E_i})
\end{equation}
where $\mathbb{1}_{\{\cdot\}}$ is the indicator function that equals 1 if the condition is satisfied and 0 otherwise, and $E_i(\cdot; \theta_{E_i})$ represents the $i$-th expert parameterized by $\theta_{E_i}$. This formulation is equivalent to activating only the selected expert $E_k$ during inference, while all remaining experts remain inactive. Consequently, the proposed hard-routing mechanism achieves model-level sparse activation and preserves the structural consistency of object detection, effectively avoiding the feature interference and computational redundancy commonly introduced by multi-expert weighted fusion in soft-routing paradigms.

\subsection{Decoupled Two-Stage Training Paradigm}\label{subsec3-4}
\label{subsec:training}

To ensure stable optimization of both the expert models and the gating network, we adopt a decoupled two-stage training paradigm. This design decomposes the joint optimization process into two sequential optimization stages, thereby alleviating the optimization instability caused by simultaneous routing and expert learning.

\paragraph{Stage I: Independent Expert Optimization.}

In the first stage, each expert model is independently trained on its corresponding domain-specific dataset $\mathcal{D}_i$ ($i \in \{C, S, W\}$). The optimal parameters $\theta_{E_i}^*$ for each expert are obtained by minimizing the detection loss without any shared parameters:
\begin{equation}
    \theta_{E_i}^* =
    \arg\min_{\theta_{E_i}}
    \mathbb{E}_{(I, b_{gt}) \sim \mathcal{D}_i}
    \left[
    \mathcal{L}_{det}\left(E_i(I; \theta_{E_i}), b_{gt}\right)
    \right]
\end{equation}
where $b_{gt}$ represents the ground-truth bounding boxes and class labels. The generic optimization objective for the detection task $\mathcal{L}_{det}$ is defined as:
\begin{equation}
    \mathcal{L}_{det}
    =
    \mathcal{L}_{cls}
    +
    \lambda_{box}\mathcal{L}_{box}
    +
    \lambda_{obj}\mathcal{L}_{obj}
\end{equation}
where $\mathcal{L}_{cls}$, $\mathcal{L}_{box}$, and $\mathcal{L}_{obj}$ denote the classification loss, bounding box regression loss, and objectness loss, respectively.

\paragraph{Stage II: Gating Network Optimization.}

In the second stage, the optimized expert parameters are frozen ($\theta_{E_i} = \theta_{E_i}^*$). The gating network is then trained as a multi-class classifier supervised by explicit scene-domain labels over the mixed dataset $\mathcal{D}_{mix}$. The optimal gating parameters $\theta_G^*$ are obtained by minimizing the cross-entropy routing loss:
\begin{equation}
    \theta_G^*
    =
    \arg\min_{\theta_G}
    \mathbb{E}_{(I,y)\sim\mathcal{D}_{mix}}
    \left[
    -
    \sum_{i \in \{C,S,W\}}
    \mathbb{1}_{\{y=i\}}
    \log(p_i)
    \right]
\end{equation}
where $y$ denotes the ground-truth scene-domain label associated with the input image $I$. During this stage, only the gating network is updated, while all expert models remain frozen. This strategy effectively avoids the routing instability and expert interference that may arise from implicit joint optimization.

During inference, the total computational complexity of the system is dynamically determined by the scene-dependent routing distribution. For a single input image $I$, the inference cost can be expressed as:
\begin{equation}
    \mathit{Cost}(I)
    =
    \mathit{Cost}(G)
    +
    \mathit{Cost}(E_k)
    \approx
    \mathit{Cost}(E_k)
\end{equation}
where $\mathit{Cost}(G)$ denotes the computational overhead of the lightweight gating network.

Furthermore, the expected computational complexity over the overall data distribution can be formulated as:
\begin{equation}
    \mathbb{E}[\mathit{Cost}] = \mathit{Cost}(G) + \sum_{i \in {C,S,W}} P(y=i)\cdot \mathit{Cost}(E_i)
\end{equation}
where $P(y=i)$ denotes the routing probability of the corresponding scene domain. Since the computational overhead of the MobileNetV3-Small gating network is negligible and a large proportion of standard driving samples are routed to the lightweight expert, the proposed sparse activation mechanism effectively reduces the average computational workload while preserving robust perception capability in challenging scenarios such as distant small-object and adverse-weather environments.

As a result, the proposed training and inference paradigm enables the framework to dynamically select the most suitable parameter subspace according to the input characteristics, thereby improving detection robustness while maintaining real-time inference performance.

\section{Experiments}\label{sec4}

\subsection{Experimental Setup}\label{subsec4-1}
\noindent\textbf{Datasets:} To validate the adaptive perception capability of the proposed method across diverse traffic scenarios, a composite multi-domain traffic sign dataset encompassing three typical driving conditions is constructed. Specifically:
\begin{enumerate}[1.]
    \item \textbf{Standard Clear Domain (C)}: The data originates from the Roboflow platform, containing 3,626, 779, and 797 images for the training, testing, and validation sets, respectively. Quantitative definition: referencing the medium and large object standards defined in COCO (Common Objects in Context), samples with a target pixel area of $\text{Area} > 32^2$ and good visibility are selected to simulate conventional perception scenarios.~\cite{ref28, ref29}
    \item \textbf{Distant Small-Object Domain (S)}: The data is sourced from the TT100K (Tsinghua-Tencent 100K) dataset, which comprises 4,746, 1,017, and 1,017 images for training, testing, and validation, respectively. Targets with a pixel area of $\text{Area} \le 32^2$ are filtered to characterize small-scale object detection challenges under distant or high-resolution perspectives.
    \item \textbf{Adverse Weather Degradation Domain (W)}: This domain is constructed by applying rain, snow, and fog augmentations to the aforementioned two categories of data. It contains 8,372, 1,814, and 1,796 images for training, testing, and validation, respectively, to simulate perception scenarios under low visibility and intense environmental interference.
\end{enumerate}
Furthermore, a mixed dataset containing 7,000, 1,500, and 1,500 images is constructed via stratified sampling according to a 5:3:2 ratio (C:S:W). All experiments uniformly adopt a 14-class traffic sign setup and employ a category encoding system consistent with TT100K to ensure evaluation consistency.

\medskip
\noindent\textbf{Evaluation Metrics:} This study utilizes the official COCO evaluation protocol to assess model performance. Specifically, mAP50 and mAP50-95 from the mean Average Precision (mAP) metrics are selected to measure detection accuracy, with mAP50-95 serving as the core index to comprehensively reflect model robustness. For efficiency evaluation, the average computational workload (Avg. GFLOPs) and frames per second (FPS) are reported. Given the dynamic routing characteristics, GFLOPs are calculated as a weighted average based on the test set distribution (5:3:2), covering the combined complexity of both the gating network and the expert models. Due to the deployment of a lightweight gating network, its computational overhead is extremely negligible (approximately 0.06 GFLOPs) and is therefore omitted from the total complexity calculation. FPS tracks the end-to-end inference latency under a fixed hardware environment, with the computation process encompassing both the gating inference and the forward propagation of the single activated expert. Notably, expert switching only involves index logic jumps within the GPU memory, introducing no additional model-loading latency, thereby faithfully reflecting the real-time operational efficiency of the system.

\medskip
\noindent\textbf{Model Settings:} The CBDES MoE TSR framework is developed based on PyTorch 2.9.1 and Ultralytics YOLO, taking only raw RGB images as input. The expert pool is composed of three heterogeneous detectors initialized with COCO pre-trained weights: YOLOv11s tailored for clear, near-range scenarios, and two separate YOLOv9c models dedicated to distant small objects and adverse weather scenarios, respectively. The gating network adopts the MobileNetV3-Small architecture to achieve lightweight, image-level expert routing.

\medskip
\noindent\textbf{Training Configuration:} The training process is divided into two distinct stages. First, all expert models undergo training for 150 epochs, with the input resolution uniformly set to $640 \times 640$ and a batch size of 8. The optimization process employs a Stochastic Gradient Descent (SGD) optimizer with an initial learning rate of 0.01, a momentum coefficient of 0.937, and a weight decay coefficient of 0.0005, supplemented by Mosaic and MixUp augmentation strategies. For Faster R-CNN, as a baseline comparison, we only load the pre-trained weights and apply a standard fine-tuning strategy to train for 30 epochs. Subsequently, in the gating network training stage, the input resolution is adjusted to $224 \times 224$ for a duration of 40 epochs. During this phase, all expert parameters are frozen, and the cross-entropy loss is optimized solely under the supervision of explicit domain labels, thereby ensuring the functional independence and decoupling of each expert.

\medskip
\noindent\textbf{Experimental Platform:} All experiments are executed on a single NVIDIA GeForce RTX 5060 Laptop GPU equipped with 8 GB of VRAM. The software environment comprises the Ubuntu 22.04 operating system, the PyTorch 2.9.1 deep learning framework, and the CUDA 12.8 acceleration library.

\subsection{Main Results}\label{subsec4-2}

As shown in Table~\ref{tab1}, CBDES MoE TSR outperforms the compared static detection methods across all evaluation metrics on the composite multi-domain traffic sign dataset. Specifically, CBDES MoE TSR achieves an mAP50-95 of 76.8\%, improving the primary metric by 2.3\% compared with the baseline YOLOv9c model trained on the mixed dataset. In addition, we compare the proposed method with the classical two-stage detector Faster R-CNN. Although Faster R-CNN is evaluated under a limited training schedule as a basic reference baseline, CBDES MoE TSR still demonstrates a clear performance advantage. These quantitative results verify that the proposed framework provides strong detection capability across diverse traffic scenarios.

\begin{table*}[h]
\caption{Performance comparison of different detection methods on the traffic sign dataset}\label{tab1}%
\begin{tabular}{@{}llcccc@{}}
\toprule
Method & Type & mAP50 (\%) & mAP50-95 (\%) & Avg. GFLOPs & FPS \\
\midrule
Faster R-CNN (ResNet50) & Two-Stage & 81.4 & 65.3 & 134.01 & 17.99 \\
YOLOv9c                 & Baseline & 88.3 & 74.5 & 103.70 & 53.78 \\
CBDES MoE TSR           & MoE       & \textbf{92.6} & \textbf{76.8} & \textbf{62.83} & \textbf{58.82} \\
\botrule
\end{tabular}

\vspace{1.5ex} 
{\footnotesize \noindent \textbf{Note:} All models use an input resolution of $640 \times 640$. Type denotes the detector architecture category. Avg. GFLOPs represents the average computational cost weighted by the test-set scenario distribution (C:S:W=5:3:2). FPS is measured on a single NVIDIA GeForce RTX 5060 Laptop GPU. Bold values indicate the best result in each column.\par}

\end{table*}

In terms of inference efficiency, the corresponding results are also reported in Table~\ref{tab1}. The average computational cost of CBDES MoE TSR is reduced by approximately 39.4\% and 53.1\% compared with YOLOv9c and Faster R-CNN, respectively. This improvement mainly benefits from the hierarchically decoupled heterogeneous expert pool, which dynamically allocates computational resources according to the semantic characteristics of the input images. As a result, the proposed framework effectively reduces the average computational cost while maintaining high detection accuracy.

\begin{figure}[htbp]
\centering
\includegraphics[width=0.95\columnwidth]{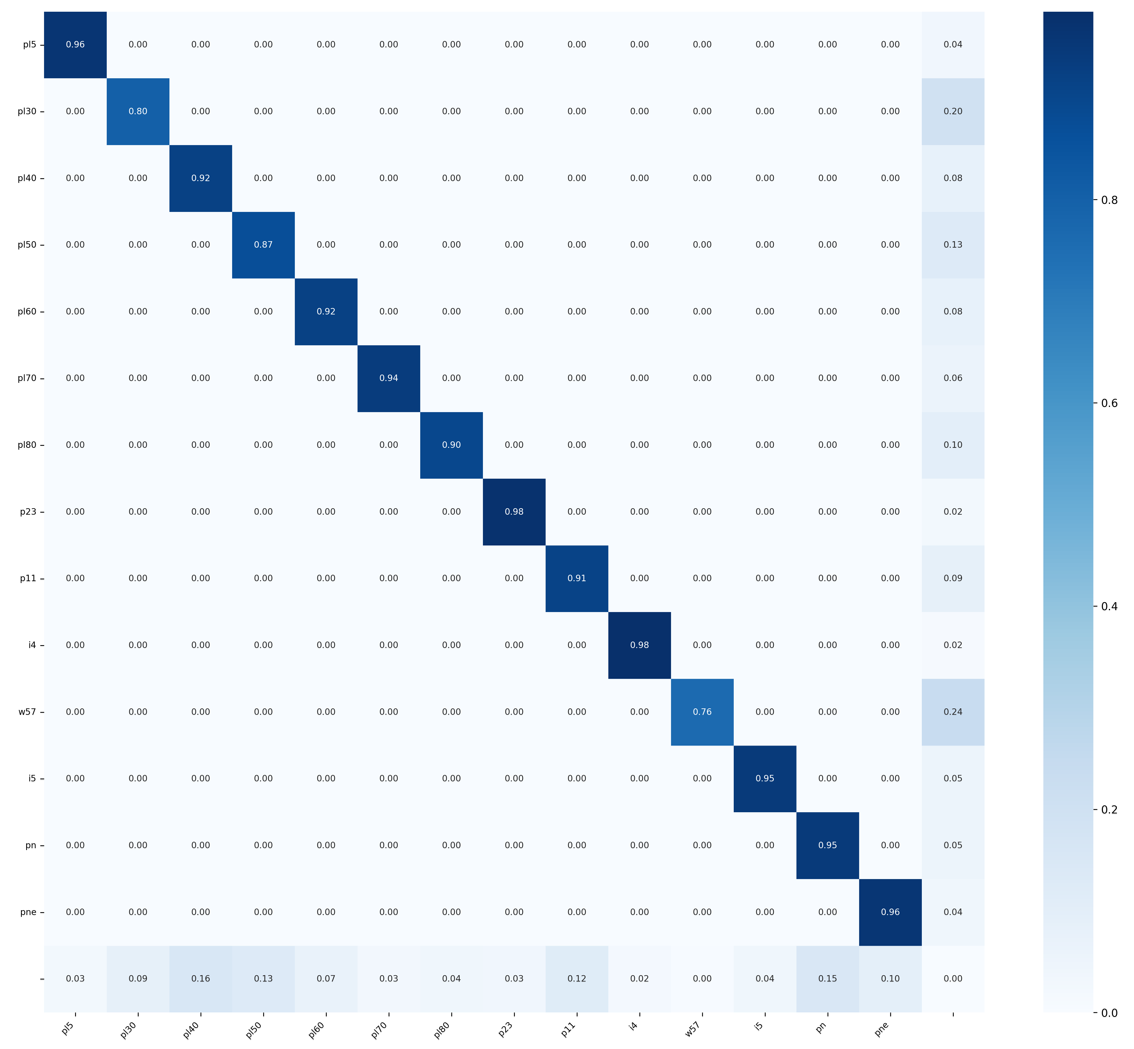} 
\caption{Confusion matrix for traffic sign detection by the CBDES MoE TSR system on the test set. The diagonal values represent the proportion of correctly detected instances for each category. The rightmost column indicates the proportion of ground-truth targets missed (predicted as background), and the bottom row shows the proportion of background instances incorrectly detected as corresponding signs. The results demonstrate that the system maintains high detection accuracy across most categories, with inter-class confusion being effectively suppressed.}\label{fig3}
\end{figure}

In addition to the macroscopic evaluation of overall detection accuracy, to further investigate the performance of the CBDES MoE TSR framework on fine-grained categories, we visualize the multi-class classification and background confusion matrix evaluated on the test set, as shown in Fig.~\ref{fig3}. As illustrated, the diagonal of the confusion matrix exhibits a high-confidence distribution, demonstrating that the proposed framework possesses robust classification and localization capabilities across the vast majority of traffic sign categories. For instance, the accuracies for the p23 and i4 categories both reach 0.98, while pne and pl5 also exceed 0.96. Notably, the rightmost column of the matrix reveals the probability of targets being misclassified as the background, representing the missed detection rate. Among these, the w57 and pl30 categories exhibit relatively higher missed detection rates of 0.24 and 0.20, respectively. This is primarily because such signs in real-world scenarios typically occupy an extremely small proportion of pixels, or are frequently obscured by visual degradation caused by adverse weather conditions. Nevertheless, owing to the heterogeneous expert collaboration mechanism and the specialized representation capabilities tailored for complex scenarios, the system successfully controls the missed detection rate of these challenging samples within a reasonable range while simultaneously maintaining a low inter-class misclassification rate. This further validates the comprehensiveness and robustness of the proposed method from the perspective of fine-grained category evaluation.

\subsection{Expert Selection Study}\label{subsec4-3}

To validate the rationality of the selected heterogeneous expert combination, we first evaluate the independent performance of different detector architectures across the target sub-domains. As a lightweight candidate expert, YOLOv11s is introduced based on its public implementation. Its computational cost is 21.60 GFLOPs, which is significantly lower than the 103.70 GFLOPs required by YOLOv9c, making it suitable for low-complexity scenarios. This model is used only to verify the replaceability of the proposed framework rather than serving as a standard comparison baseline.

The evaluation results are presented in Table~\ref{tab2}. In the Test-C domain representing standard clear scenes, the mixed-trained YOLOv11s achieves a good balance between accuracy and efficiency, reaching 93.1\% mAP50--95. In the Test-S domain dominated by small-object scenarios, the single YOLOv9c expert trained specifically on the small-object dataset achieves 64.6\% mAP50--95, outperforming the general baseline. In the Test-W domain representing adverse-weather conditions, the YOLOv9c expert trained specifically on the adverse-weather dataset exhibits more stable performance. Based on these independent evaluations, we determine the final expert candidate configuration used in the proposed framework.

\begin{table*}[h]
\caption{Performance of different detector variants across scenario-specific test domains}\label{tab2}%
\begin{tabular}{@{}lllcccc@{}}
\toprule
Domain & Model variant & Training set & Role & mAP50(\%) & mAP50-95(\%) & Gain(\%) \\
\midrule
\multirow{3}{*}{Test-C} & YOLOv9c  & Mixed-trained       & reference  & 98.5 & 93.7 & - \\
                        & YOLOv11s & Domain-specific (C) & comparison & 98.4 & 90.8 & -0.1 \\
                        & YOLOv11s & Mixed-trained       & selected   & 98.5 & 93.1 & - \\
\midrule
\multirow{3}{*}{Test-S} & YOLOv9c  & Mixed-trained       & reference  & 76.1 & 56.2 & - \\
                        & YOLOv11s & Mixed-trained       & comparison & 67.0 & 47.9 & -9.1 \\
                        & YOLOv9c  & Domain-specific (S) & selected   & \textbf{86.6} & \textbf{64.6} & \textbf{+10.5} \\
\midrule
\multirow{3}{*}{Test-W} & YOLOv9c  & Mixed-trained       & reference  & 84.2 & 67.7 & - \\
                        & YOLOv11s & Mixed-trained       & comparison & 78.0 & 62.9 & -6.2 \\
                        & YOLOv9c  & Domain-specific (W) & selected   & \textbf{88.9} & \textbf{67.6} & \textbf{+4.7} \\
\botrule
\end{tabular}

\vspace{1.5ex}
{\footnotesize \noindent \textbf{Note:} Domain indicates the evaluation subset (C: clear scene, S: small-object scenario, W: adverse weather). Mixed-trained denotes models trained on the combined dataset, while domain-specific refers to models trained only on the corresponding sub-domain. Selected indicates the expert configuration adopted in the final MoE framework. Gain represents the change relative to the mixed-trained YOLOv9c baseline on mAP50.\par}

\end{table*}

\begin{figure*}[htbp]
\centering
\includegraphics[width=0.95\textwidth]{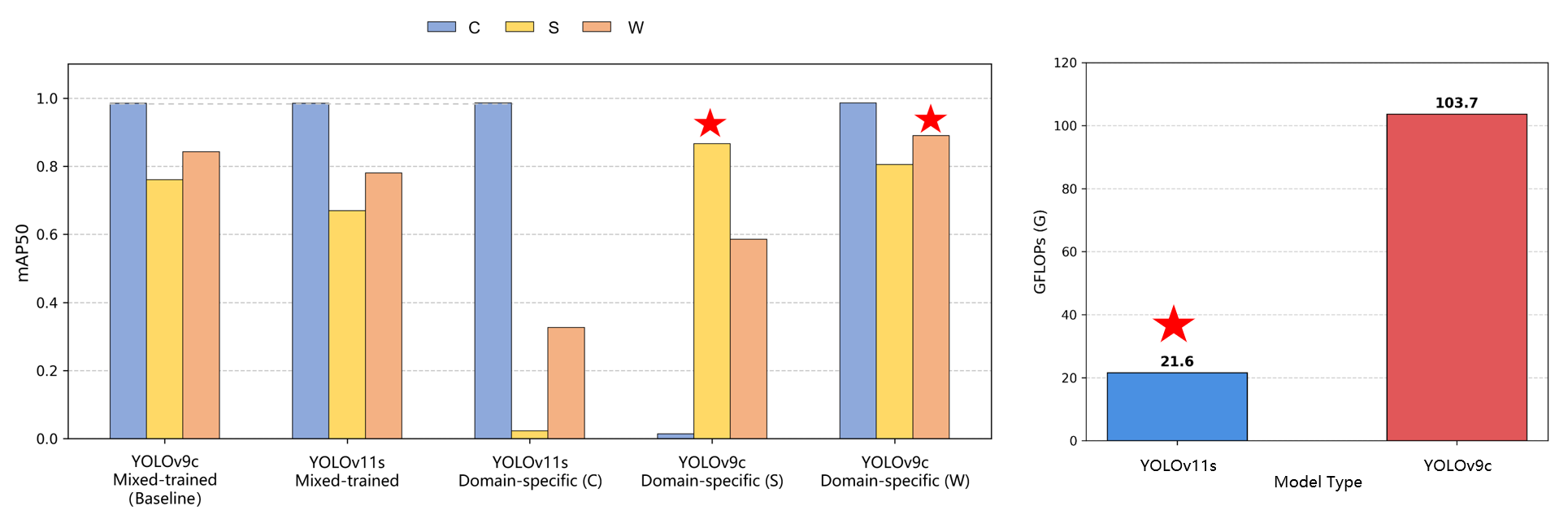} 
\caption{Visual analysis of the rationality of expert pool construction. The left plot displays the mAP50 performance comparison of individual models under different training strategies across various traffic scenario sub-domains, where red stars mark the specialized expert models that exhibit optimal performance in specific complex scenarios (S and W domains). The right plot compares the single-inference computational cost (Avg. GFLOPs) of candidate expert models, with the red star highlighting the prominent lightweight advantage of YOLOv11s. Combining both plots, it is evident that the heterogeneous expert combination selected in this study secures peak performance across all sub-domains while possessing an extremely low computational lower bound.}\label{fig4}
\end{figure*}

These results indicate that different training data distributions lead to distinct feature biases in the learned models. Mixed training improves the model's generalization ability across multiple scenarios, whereas sub-domain-specific training strengthens the representational capability for particular scenarios. Further analysis shows that in the Test-C domains, models trained on the mixed dataset generally outperform or approach the performance of those trained on individual sub-domains. In contrast, in the Test-S domain dominated by small-object scenarios, the specialized model still exhibits a clear advantage, indicating that dedicated training for distant small-scale targets is necessary. To eliminate the influence of data-scale differences, we conduct a controlled experiment using the same amount of small-object training data. The results show that, even under identical data-scale conditions, the small-object specialized model still achieves higher mAP than the mixed-trained model. This observation further confirms the conclusion above.
These findings suggest that there is no single training strategy that can achieve optimal performance across all traffic scenarios. Mixed-trained improves overall robustness but weakens scenario-specific specialization, whereas sub-domain training enhances performance in particular scenarios but may reduce cross-domain stability. This inherent cross-domain performance conflict caused by a fixed parameter space provides the primary motivation for introducing the proposed dynamic expert routing mechanism. 

To more intuitively elucidate the rationale behind the selection of heterogeneous experts, the evaluation results of the aforementioned strategies and the computational costs of the models are visualized in Fig.~\ref{fig4}. As shown on the left side of Fig.~\ref{fig4}, the YOLOv9c trained on specific sub-domains demonstrates favorable detection accuracy in both small-object (S) and adverse-weather (W) scenarios, confirming the necessity of allocating high-capacity specialized experts to complex scenarios. Meanwhile, as illustrated on the right side of Fig.~\ref{fig4}, the single forward inference computational workload of the lightweight YOLOv11s model is merely 21.60~GFLOPs, which is lower than the 103.70~GFLOPs of YOLOv9c. Since YOLOv11s maintains comparable performance to the high-computation baseline in the standard clear scenario (C), selecting it as the expert for this domain effectively eliminates the redundant computational overhead induced by simple scenarios. Combining these quantitative and visual analyses, the heterogeneous expert pool constructed in this study logically achieves an optimal balance between scenario-specific specialization and overall computational efficiency.

\subsection{Framework Ablation Study}\label{subsec4-4}

Finally, to verify the stability of the CBDES MoE TSR framework, we compare the performance of several expert combination strategies under this framework, and the results are presented in Table~\ref{tab3}. Combination A represents a high-capacity expert configuration composed of three YOLOv9c detectors, including a mixed-trained YOLOv9c serving as the expert for the C domain, a small-object specialized YOLOv9c for the S domain, and an adverse-weather specialized YOLOv9c for the W domain, which serves to approximate the performance upper bound. Combination B introduces a lightweight detector, with its expert composition comprising a clear-domain specialized lightweight YOLOv11s, a small-object specialized YOLOv9c, and a mixed-trained YOLOv9c.

\begin{table*}[h]
\caption{Performance and computational efficiency of different expert configurations in the CBDES MoE TSR framework}\label{tab3}%
\begin{tabular}{@{}lcccc@{}}
\toprule
Configuration & mAP50 (\%) & mAP50-95 (\%) & Avg. GFLOPs & Gain (\%) \\
\midrule
Baseline (YOLOv9c)        & 88.3 & 74.5 & 103.70 & - \\
CBDES MoE TSR (Config-A)  & 91.7 & \textbf{77.2} & 103.70 & +3.4 \\
CBDES MoE TSR (Config-B)  & 91.6 & 75.9 & \textbf{62.83} & +3.3 \\
CBDES MoE TSR (Ours)      & \textbf{92.6} & 76.8 & \textbf{62.83} & \textbf{+4.3} \\
\botrule
\end{tabular}

\vspace{1.5ex}
{\footnotesize \noindent \textbf{Note:} Baseline denotes the single YOLOv9c baseline model trained on the mixed dataset. The first MoE row (Config-A) represents a high-capacity configuration composed of three YOLOv9c models to approximate the performance upper bound. Other configurations alter the training data distribution of experts to validate the impact of dynamic routing on performance and computational efficiency. The results show that stable accuracy improvements can be achieved across different expert compositions, while our final configuration significantly reduces the computational workload while maintaining a performance close to the upper bound. Gain represents the change relative to the baseline single model on mAP50.\par}

\end{table*}

Compared with the single YOLOv9c baseline trained on the mixed dataset, all configurations built upon the proposed framework achieve consistent accuracy improvements. Furthermore, it can be observed that although Combinations A and B possess different internal expert compositions, both consistently outperform the single-model baseline. This indicates that the performance improvement does not depend on a specific expert architecture but stems from the scenario-adaptive functional allocation capability of the framework, i.e., the model can select an appropriate parameter subspace for inference based on the characteristics of the input, thereby forming stable cross-scenario performance enhancements.

On this basis, we select the expert combination consisting of the mixed-trained YOLOv11s, the small-object sub-domain trained YOLOv9c, and the adverse-weather sub-domain data trained YOLOv9c as the default configuration (Ours). While maintaining an accuracy comparable to the full high-capacity combination, this configuration reduces the average computational workload from 103.70~GFLOPs to 62.83~GFLOPs, achieving an effective trade-off between detection accuracy and computational overhead.

\begin{figure}[htbp]
\centering
\includegraphics[width=0.95\columnwidth]{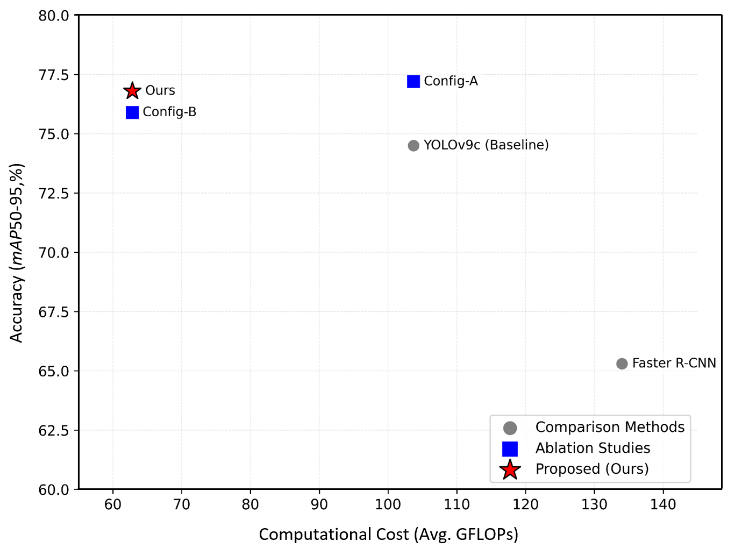}
\caption{Visual analysis of the rationality of expert pool construction. The left plot displays the mAP50 performance comparison of individual models under different training strategies across various traffic scenario sub-domains, where red stars mark the specialized expert models that exhibit optimal performance in specific complex scenarios (S and W domains). The right plot compares the single-inference computational cost (Avg. GFLOPs) of candidate expert models, with the red star highlighting the prominent lightweight advantage of YOLOv11s. Combining both plots, it is evident that the heterogeneous expert combination selected in this study secures peak performance across all sub-domains while possessing an extremely low computational lower bound.}\label{fig5}
\end{figure}

To intuitively illustrate the efficiency-performance relationship among different configurations, the results are visualized in Fig.~\ref{fig5}. As can be seen, CBDES MoE TSR lies to the left region of the high-computation baseline, indicating that it significantly reduces the average computational complexity while maintaining the overall detection performance.

\subsubsection{Gating Classification Accuracy and Confusion Matrix}\label{subsubsec4-4-1}

\begin{figure}[htbp]
\centering
\includegraphics[width=0.7\columnwidth]{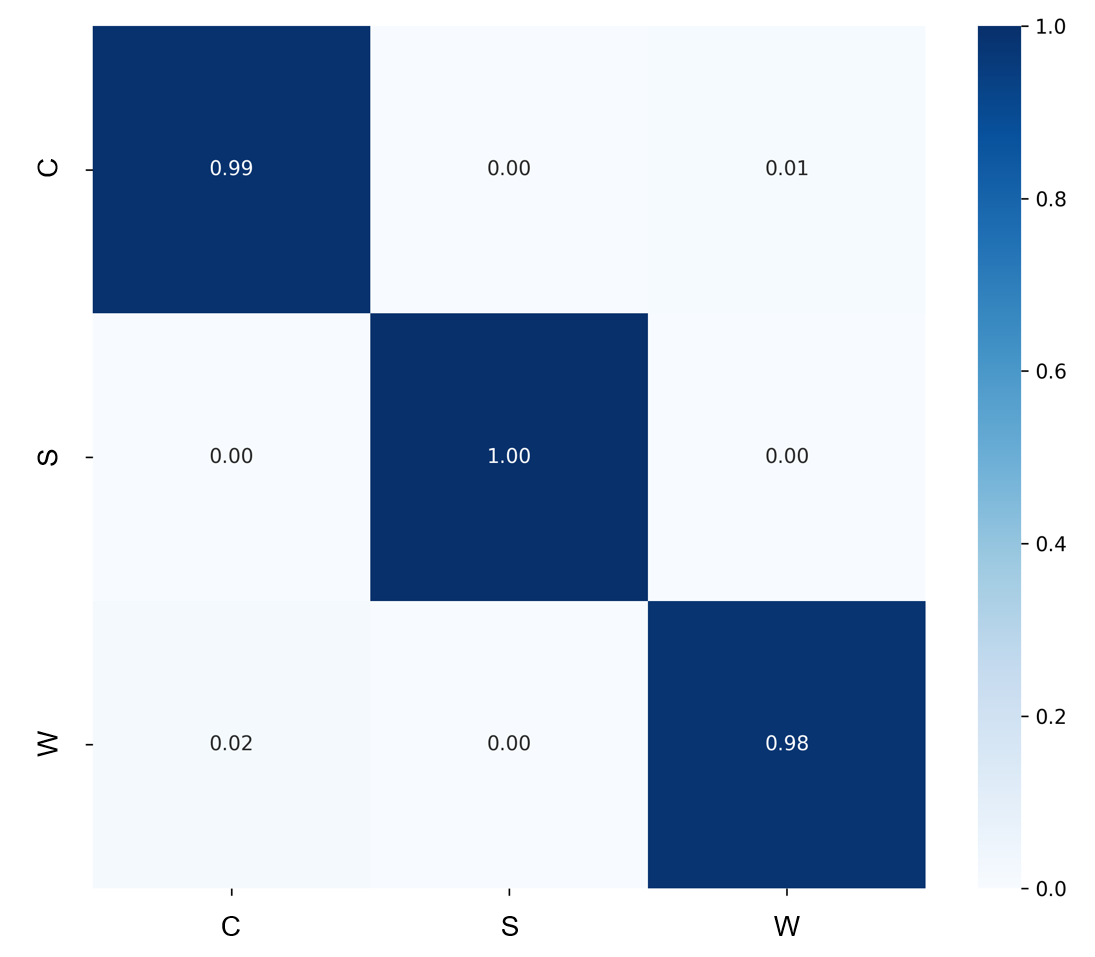} 
\caption{Confusion matrix of the gating network evaluated on the test set. The horizontal axis represents the predicted scene domain, and the vertical axis represents the ground-truth scene domain. The results show that the prediction accuracies of the gating network across all three sub-domains exceed 98\%, validating the high reliability of route allocation under explicit supervision.}\label{fig6}
\end{figure}

The decision quality of the gating network determines whether input samples can be allocated to the optimal feature subspace. To quantify the performance of the gating network, we plot the confusion matrix of scene domain classification evaluated on the test set, as shown in Fig.~\ref{fig6}. The results indicate that, benefiting from the decoupled two-stage training paradigm and explicit domain label supervision, the lightweight MobileNetV3 gating network demonstrates a strong scene discrimination capability, achieving an overall gating classification accuracy of 99.2\%. Specifically, its classification accuracies on the three sub-domains---standard clear (C), small object (S), and adverse weather (W)---reach as high as 99\%, 100\%, and 98\%, respectively. This proves that the lightweight gating network can accurately capture the global semantics and environmental degradation features of the images, thereby providing highly reliable control signals for subsequent expert routing, ensuring that the gating module does not constitute a performance bottleneck for the overall system.

\subsubsection{Routing Consistency Analysis}\label{subsubsec4-4-2}

\begin{figure*}[htbp]
\centering
\includegraphics[width=0.95\textwidth]{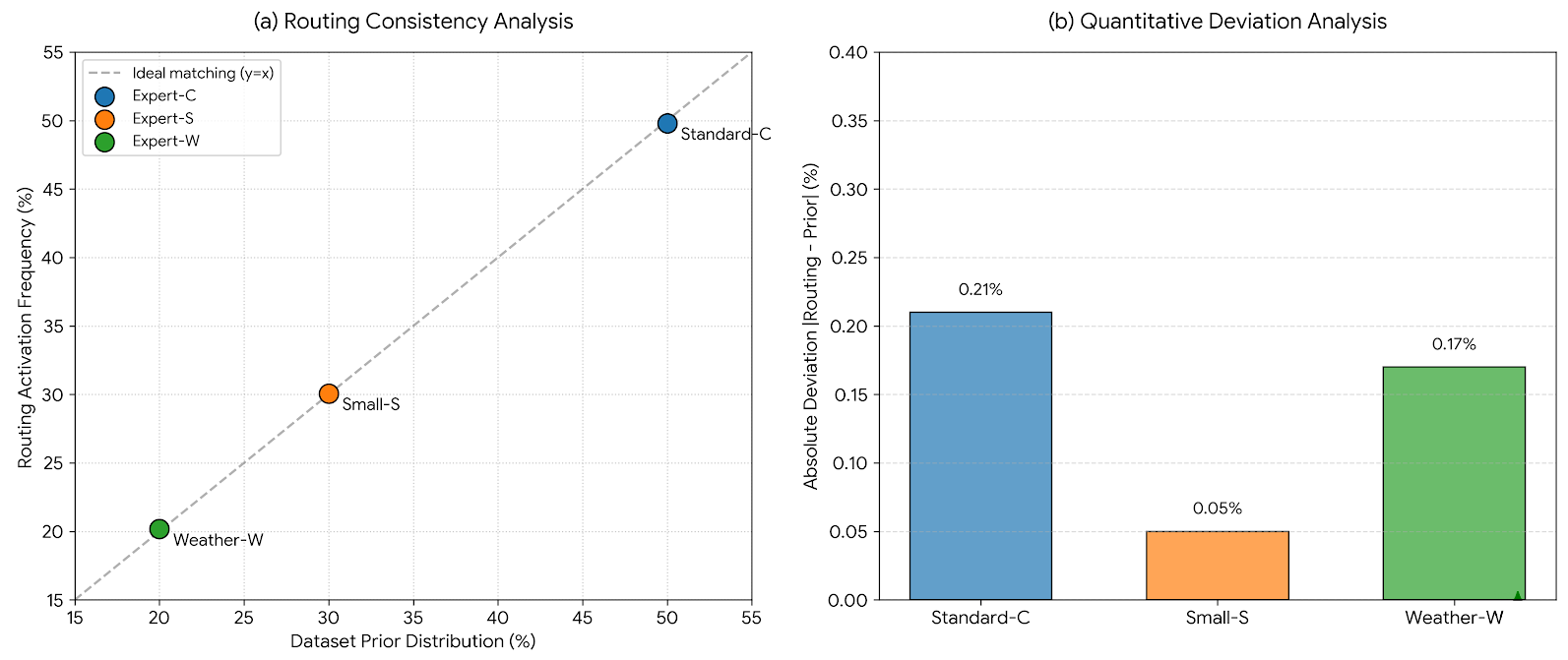}
\caption{Routing consistency analysis of the gating network. (a) The correspondence between expert activation frequency and the prior data distribution, where the dashed line represents ideal consistency ($y=x$). (b) The absolute deviation of the routing distribution for each scenario relative to the data distribution. The statistical results indicate that the routing activation distribution is highly consistent with the inherent distribution of the dataset, proving that the gating network successfully achieves a rational allocation of computational resources while effectively avoiding mode collapse.}\label{fig7}
\end{figure*}

Based on the aforementioned classification accuracy, we further analyze the actual route allocation behavior of the gating network across the entire test set and compare it with the prior scene distribution of the dataset (C:S:W=5:3:2). The results are illustrated in Fig.~\ref{fig7}. In Fig.~\ref{fig7}(a), all scene points are distributed closely along the diagonal, indicating that the expert selection of the model is generally consistent with the data distribution. The absolute deviation analysis presented in Fig.~\ref{fig7}(b) further demonstrates that the errors for all three scene categories are maintained within an extremely small range (with a maximum absolute deviation of only 0.21\%), and no obvious expert bias or mode collapse phenomenon is observed.

Specifically, approximately 49.8\% of the samples are routed to the lightweight expert, Expert-C (implemented via YOLOv11s), to process standard clear scenarios, thereby significantly reducing the average computational overhead while guaranteeing detection performance. The remaining samples are precisely allocated to experts with stronger computational capacities, among which the small-object scenarios (Expert-S) and adverse-weather scenarios (Expert-W) account for 30.05\% and 20.17\%, respectively. This on-demand scheduling strategy, which highly aligns with the inherent distribution of the data, ensures that computational resources are primarily concentrated on highly challenging and complex samples, thus mechanistically explaining the rationale behind the framework's overall efficiency improvement.

\subsubsection{Oracle Routing Upper Bound Analysis}\label{subsubsec4-4-3}

To evaluate the final performance gap between the current gating network and ``perfect routing,'' we introduce the Oracle Routing mechanism as a reference baseline for the performance upper bound. Under the Oracle setting, we mask the predictive output of the gating network and directly utilize the ground-truth scene domain labels of the test set to forcibly activate the corresponding truly matched expert for each sample.

Experimental measurements show that under the Oracle strategy, the system achieves an mAP50 of 93.5\% and an mAP50-95 of 77.2\%. In contrast, the actual performance achieved by our proposed gating network (Gating) is 92.6\% for mAP50 and 76.8\% for mAP50-95. The gap between the core metrics is less than 1\%, which further confirms that the current gating network effectively accomplishes the sample scheduling task, closely approaching the theoretical performance upper bound under this heterogeneous expert pool configuration.

\subsubsection{Noise Interference and Robustness Verification}\label{subsubsec4-4-4}

\begin{figure}[htbp]
\centering
\includegraphics[width=0.95\columnwidth]{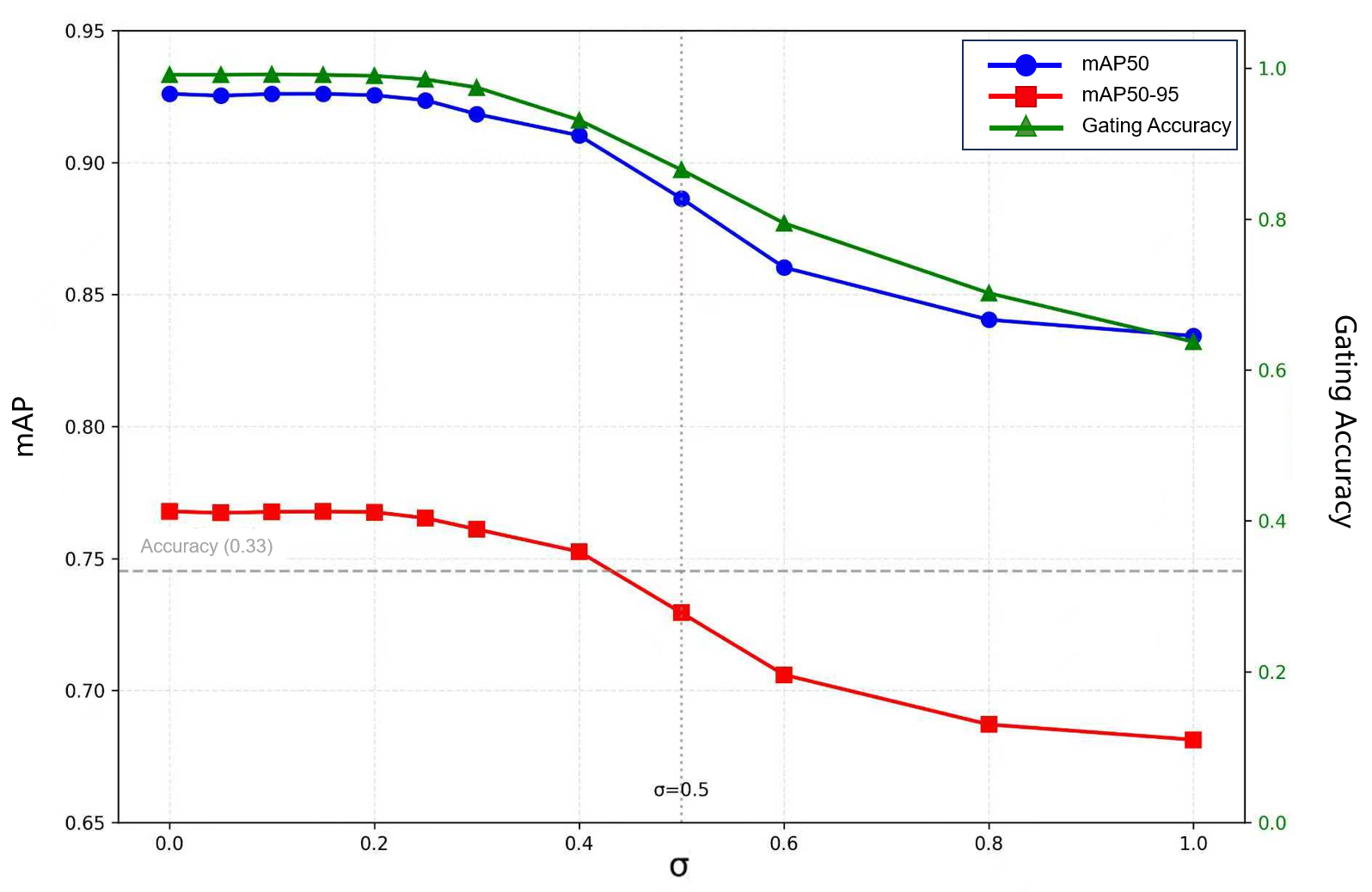} 
\caption{Noise interference and robustness analysis. The horizontal axis represents the intensity $\sigma$ of the Gaussian noise injected into the gating output. The green line indicates the gating classification accuracy. The blue and red lines represent the mAP50 and mAP50-95 of the system, respectively. As the increasing noise ($\sigma$ from 0.0 to 1.0) leads to a higher probability of misrouting, the system detection accuracy exhibits a clear downward trend, conversely proving the critical role of correctly matching experts in maintaining framework performance.}\label{fig8}
\end{figure}

To further quantitatively evaluate the system's dependence on precise routing matching and explore the negative impact of expert allocation deviation on overall detection performance, we design a gating noise interference experiment. During the inference phase, we artificially inject Gaussian noise with different standard deviations $\sigma$ into the output logits of the gating network to gradually interfere with its Top-1 decision. Fig.~\ref{fig8} illustrates the quantitative variation trends of the gating accuracy and system detection accuracy as the noise intensity increases. The experimental data clearly demonstrates the high dependence of the system's detection capability on correct routing:

\begin{itemize}
    \item \textbf{Noise-free state ($\sigma=0.0$):} The system operates in an ideal state with a gating accuracy of 99.2\%, while the overall system achieves an mAP50 and mAP50-95 of 92.6\% and 76.8\%, respectively.
    \item \textbf{Moderate interference ($\sigma=0.5$):} The gating accuracy drops to 86.5\%. Because some samples are routed to incorrect expert networks, the mAP50 of the system falls to 88.6\%, and the core robustness metric mAP50-95 decreases to 72.9\%.
    \item \textbf{Severe interference ($\sigma=1.0$):} The gating accuracy further plummets to 63.7\%. At this point, the system performance experiences a precipitous decline, with the mAP50 remaining at only 83.4\% and the mAP50-95 drastically degrading to 68.1\%.
\end{itemize}

This phenomenon compellingly proves two key conclusions. First, there is explicit functional decoupling among the heterogeneous experts; the lightweight expert lacks the capability to handle small objects, while the adverse-weather expert processing clear samples may trigger feature misalignment, indicating that incorrect routing matches can disrupt the original feature representations. Second, the dynamic routing mechanism proposed in this study is the core driving force for the framework to achieve high accuracy. Precise ``on-demand allocation'' is not only a means to improve efficiency but also an essential condition for maintaining high accuracy in complex scenarios.

\subsection{Qualitative Results}\label{subsec4-5}

\begin{figure*}[htbp]
\centering
\includegraphics[width=0.95\textwidth]{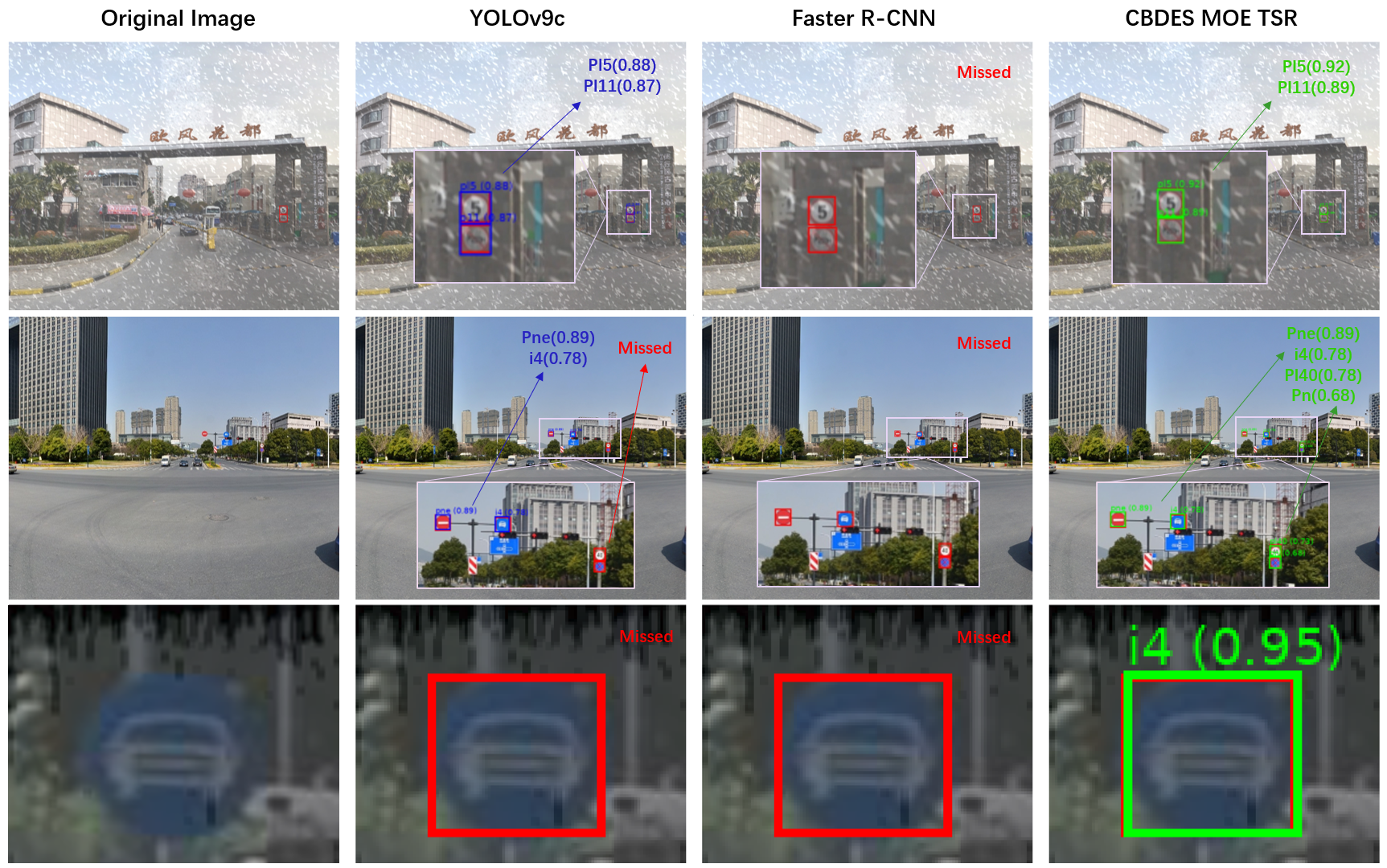} 
\caption{Qualitative comparison of detection results under different traffic scenarios. From left to right: input image, YOLOv9c, Faster R-CNN, and the proposed CBDES MoE TSR. Light-purple boxes indicate enlarged regions for visualization. Red boxes denote ground-truth annotations, blue boxes represent predictions from YOLOv9c, and green boxes correspond to predictions from CBDES MoE TSR.}\label{fig9}
\end{figure*}

To visually verify the effectiveness of the dynamic expert selection mechanism, we present the detection results of a single detector and our proposed method across different scenarios. As illustrated in Fig.~\ref{fig9}, the static single model exhibits clear limitations in adaptability across cross-domain scenarios, frequently resulting in missed detections or low-confidence predictions under distant small-object and adverse-weather conditions. In contrast, after CBDES MoE TSR routes the samples to the matching experts through the gating network, the detection performance is significantly improved. This qualitative comparison successfully validates the necessity of input-aware dynamic inference in complex traffic scenarios.

\section{Conclusion}\label{sec5}

This paper addresses the core challenge of insufficient input adaptability faced by static single-stage detectors in complex traffic sign perception tasks. Because conventional detectors employ fixed parameters and computational paths during inference, existing methods struggle to achieve adaptive representation in multi-domain traffic scenarios. To tackle this issue, we propose CBDES MoE TSR, a heterogeneous mixture-of-experts detection framework based on the concept of hierarchical decoupling. By constructing an expert pool composed of heterogeneous YOLO detectors and introducing an explicitly supervised image-level gating mechanism, CBDES MoE TSR can dynamically select the matching expert model according to the input scene. This enhances representation adaptability for specific scenarios such as small targets while keeping the inference structure concise and controllable, successfully realizing a paradigm shift from ``fixed-parameter fitting'' to ``dynamic on-demand representation.'' Experimental results on the composite multi-domain traffic sign dataset demonstrate that, compared with the single high-capacity detection model, CBDES MoE TSR improves the detection accuracy by 2.3\% in mAP50-95 over the YOLOv9c baseline, while reducing the average computational workload by approximately 39.4\%. These results validate the effectiveness of image-level hard routing and heterogeneous expert collaboration in traffic sign detection tasks.

Regarding limitations, the proposed method relies on explicit domain partitioning of the training data and gating supervision, which to some extent increases the complexity of data preparation and system design. Furthermore, although the proposed explicit routing strategy improves controllability and deployment efficiency, complex hybrid scenarios involving simultaneous small-object and adverse-weather degradation may still introduce partial domain overlap between experts. In addition, the current gating strategy adopts a static offline training paradigm and has not yet been adaptively optimized for distribution shifts or long-term operational scenarios. Future work will investigate adaptive multi-condition routing, finer-grained expert decomposition strategies, and lightweight expert designs, while further exploring the potential of online and continual learning to improve system robustness and cross-scenario generalization capability.

\backmatter

\bmhead{Supplementary information}

Not applicable.

\bmhead{Acknowledgements}

This work was supported by the National Key R\&D Program of China under Grant No. 2024YFB2505500.





\begin{thebibliography}{29}
\ifx \bisbn   \undefined \def \bisbn  #1{ISBN #1}\fi
\ifx \binits  \undefined \def \binits#1{#1}\fi
\ifx \bauthor  \undefined \def \bauthor#1{#1}\fi
\ifx \batitle  \undefined \def \batitle#1{#1}\fi
\ifx \bjtitle  \undefined \def \bjtitle#1{#1}\fi
\ifx \bvolume  \undefined \def \bvolume#1{\textbf{#1}}\fi
\ifx \byear  \undefined \def \byear#1{#1}\fi
\ifx \bissue  \undefined \def \bissue#1{#1}\fi
\ifx \bfpage  \undefined \def \bfpage#1{#1}\fi
\ifx \blpage  \undefined \def \blpage #1{#1}\fi
\ifx \burl  \undefined \def \burl#1{\textsf{#1}}\fi
\ifx \doiurl  \undefined \def \doiurl#1{\url{https://doi.org/#1}}\fi
\ifx \betal  \undefined \def \betal{\textit{et al.}}\fi
\ifx \binstitute  \undefined \def \binstitute#1{#1}\fi
\ifx \binstitutionaled  \undefined \def \binstitutionaled#1{#1}\fi
\ifx \bctitle  \undefined \def \bctitle#1{#1}\fi
\ifx \beditor  \undefined \def \beditor#1{#1}\fi
\ifx \bpublisher  \undefined \def \bpublisher#1{#1}\fi
\ifx \bbtitle  \undefined \def \bbtitle#1{#1}\fi
\ifx \bedition  \undefined \def \bedition#1{#1}\fi
\ifx \bseriesno  \undefined \def \bseriesno#1{#1}\fi
\ifx \blocation  \undefined \def \blocation#1{#1}\fi
\ifx \bsertitle  \undefined \def \bsertitle#1{#1}\fi
\ifx \bsnm \undefined \def \bsnm#1{#1}\fi
\ifx \bsuffix \undefined \def \bsuffix#1{#1}\fi
\ifx \bparticle \undefined \def \bparticle#1{#1}\fi
\ifx \barticle \undefined \def \barticle#1{#1}\fi
\bibcommenthead
\ifx \bconfdate \undefined \def \bconfdate #1{#1}\fi
\ifx \botherref \undefined \def \botherref #1{#1}\fi
\ifx \url \undefined \def \url#1{\textsf{#1}}\fi
\ifx \bchapter \undefined \def \bchapter#1{#1}\fi
\ifx \bbook \undefined \def \bbook#1{#1}\fi
\ifx \bcomment \undefined \def \bcomment#1{#1}\fi
\ifx \oauthor \undefined \def \oauthor#1{#1}\fi
\ifx \citeauthoryear \undefined \def \citeauthoryear#1{#1}\fi
\ifx \endbibitem  \undefined \def \endbibitem {}\fi
\ifx \bconflocation  \undefined \def \bconflocation#1{#1}\fi
\ifx \arxivurl  \undefined \def \arxivurl#1{\textsf{#1}}\fi
\csname PreBibitemsHook\endcsname

\bibitem[\protect\citeauthoryear{Chen et~al.}{2024}]{ref1}
\begin{barticle}
\bauthor{\bsnm{Chen}, \binits{H.}},
\bauthor{\bsnm{Zhang}, \binits{L.}},
\bauthor{\bsnm{Wang}, \binits{Y.}}:
\batitle{Computational methods for automatic traffic signs detection and recognition: A review}.
\bjtitle{Array}
\bvolume{23--24},
\bfpage{100331}
(\byear{2024})
\end{barticle}
\endbibitem

\bibitem[\protect\citeauthoryear{Wang et~al.}{2023}]{ref2}
\begin{bchapter}
\bauthor{\bsnm{Wang}, \binits{C.-Y.}},
\bauthor{\bsnm{Bochkovskiy}, \binits{A.}},
\bauthor{\bsnm{Liao}, \binits{H.-Y.M.}}:
\bctitle{{YOLOv7}: Trainable bag-of-freebies sets new state-of-the-art for real-time object detectors}.
In: \bbtitle{Proceedings of the IEEE/CVF Conference on Computer Vision and Pattern Recognition (CVPR)}
(\byear{2023})
\end{bchapter}
\endbibitem

\bibitem[\protect\citeauthoryear{Liu et~al.}{2016}]{ref3}
\begin{bchapter}
\bauthor{\bsnm{Liu}, \binits{W.}}, \betal:
\bctitle{{SSD}: Single shot multibox detector}.
In: \bbtitle{European Conference on Computer Vision (ECCV)}
(\byear{2016}).
\bcomment{Springer}
\end{bchapter}
\endbibitem

\bibitem[\protect\citeauthoryear{Zou et~al.}{2023}]{ref4}
\begin{barticle}
\bauthor{\bsnm{Zou}, \binits{Z.}},
\bauthor{\bsnm{Chen}, \binits{K.}},
\bauthor{\bsnm{Shi}, \binits{Z.}},
\bauthor{\bsnm{Guo}, \binits{Y.}},
\bauthor{\bsnm{Ye}, \binits{J.}}:
\batitle{Object detection in 20 years: A survey}.
\bjtitle{Proceedings of the IEEE}
\bvolume{111}(\bissue{3}),
\bfpage{257}--\blpage{276}
(\byear{2023})
\end{barticle}
\endbibitem

\bibitem[\protect\citeauthoryear{Han et~al.}{2022}]{ref5}
\begin{barticle}
\bauthor{\bsnm{Han}, \binits{Y.}}, \betal:
\batitle{Dynamic neural networks: A survey}.
\bjtitle{IEEE Transactions on Pattern Analysis and Machine Intelligence (TPAMI)}
\bvolume{44}(\bissue{11}),
\bfpage{7436}--\blpage{7456}
(\byear{2022})
\end{barticle}
\endbibitem

\bibitem[\protect\citeauthoryear{Michaelis et~al.}{2021}]{ref6}
\begin{bchapter}
\bauthor{\bsnm{Michaelis}, \binits{C.}}, \betal:
\bctitle{Benchmarking robustness in object detection: Autonomous driving when winter is coming}.
In: \bbtitle{International Conference on Learning Representations (ICLR)}
(\byear{2021})
\end{bchapter}
\endbibitem

\bibitem[\protect\citeauthoryear{Shazeer et~al.}{2017}]{ref7}
\begin{bchapter}
\bauthor{\bsnm{Shazeer}, \binits{N.}}, \betal:
\bctitle{Outrageously large neural networks: The sparsely-gated mixture-of-experts layer}.
In: \bbtitle{International Conference on Learning Representations (ICLR)}
(\byear{2017})
\end{bchapter}
\endbibitem

\bibitem[\protect\citeauthoryear{Riquelme et~al.}{2021}]{ref8}
\begin{bchapter}
\bauthor{\bsnm{Riquelme}, \binits{C.}}, \betal:
\bctitle{Scaling vision with sparse mixture of experts}.
In: \bbtitle{Advances in Neural Information Processing Systems (NeurIPS)}
(\byear{2021})
\end{bchapter}
\endbibitem

\bibitem[\protect\citeauthoryear{Wang et~al.}{2018}]{ref9}
\begin{bchapter}
\bauthor{\bsnm{Wang}, \binits{X.}}, \betal:
\bctitle{{SkipNet}: Learning dynamic routing in convolutional networks}.
In: \bbtitle{European Conference on Computer Vision (ECCV)}
(\byear{2018})
\end{bchapter}
\endbibitem

\bibitem[\protect\citeauthoryear{Gao et~al.}{2019}]{ref10}
\begin{bchapter}
\bauthor{\bsnm{Gao}, \binits{X.}}, \betal:
\bctitle{Dynamic channel pruning: Feature boosting and suppression}.
In: \bbtitle{International Conference on Learning Representations (ICLR)}
(\byear{2019})
\end{bchapter}
\endbibitem

\bibitem[\protect\citeauthoryear{Puigcerver et~al.}{2023}]{ref11}
\begin{bchapter}
\bauthor{\bsnm{Puigcerver}, \binits{J.}}, \betal:
\bctitle{From sparse to soft mixture of experts}.
In: \bbtitle{International Conference on Computer Vision (ICCV)}
(\byear{2023})
\end{bchapter}
\endbibitem

\bibitem[\protect\citeauthoryear{Xiang et~al.}{2025}]{ref12}
\begin{botherref}
\oauthor{\bsnm{Xiang}, \binits{Q.}},
\oauthor{\bsnm{Shi}, \binits{K.}},
\oauthor{\bsnm{Lin}, \binits{Z.}},
\oauthor{\bsnm{He}, \binits{L.}}:
{CBDES MoE}: Hierarchically decoupled mixture-of-experts for functional modules in autonomous driving.
arXiv preprint arXiv:2508.07838
(2025)
\end{botherref}
\endbibitem

\bibitem[\protect\citeauthoryear{Carranza-Garc{\'\i}a et~al.}{2021}]{ref13}
\begin{barticle}
\bauthor{\bsnm{Carranza-Garc{\'\i}a}, \binits{M.}}, \betal:
\batitle{On the performance of one-stage and two-stage object detectors in autonomous vehicles using camera data}.
\bjtitle{Remote Sensing}
\bvolume{13}(\bissue{1}),
\bfpage{89}
(\byear{2021})
\end{barticle}
\endbibitem

\bibitem[\protect\citeauthoryear{Wang et~al.}{2024a}]{ref14}
\begin{botherref}
\oauthor{\bsnm{Wang}, \binits{C.-Y.}},
\oauthor{\bsnm{Yeh}, \binits{I.-H.}},
\oauthor{\bsnm{Liao}, \binits{H.-Y.M.}}:
{YOLOv9}: Learning what you want to learn using programmable gradient information.
arXiv preprint arXiv:2402.13616
(2024)
\end{botherref}
\endbibitem

\bibitem[\protect\citeauthoryear{Wang et~al.}{2024b}]{ref15}
\begin{botherref}
\oauthor{\bsnm{Wang}, \binits{A.}}, et al.:
{YOLOv10}: Real-time end-to-end object detection.
arXiv preprint arXiv:2405.14458
(2024)
\end{botherref}
\endbibitem

\bibitem[\protect\citeauthoryear{Huang et~al.}{2018}]{ref16}
\begin{bchapter}
\bauthor{\bsnm{Huang}, \binits{G.}}, \betal:
\bctitle{Multi-scale dense networks for resource efficient image classification}.
In: \bbtitle{International Conference on Learning Representations (ICLR)}
(\byear{2018})
\end{bchapter}
\endbibitem

\bibitem[\protect\citeauthoryear{Liu and Deng}{2018}]{ref17}
\begin{bchapter}
\bauthor{\bsnm{Liu}, \binits{L.}},
\bauthor{\bsnm{Deng}, \binits{J.}}:
\bctitle{Dynamic deep neural networks: Optimizing accuracy--efficiency trade-offs by selective execution}.
In: \bbtitle{Proceedings of the AAAI Conference on Artificial Intelligence}
(\byear{2018})
\end{bchapter}
\endbibitem

\bibitem[\protect\citeauthoryear{Lepikhin et~al.}{2021}]{ref18}
\begin{bchapter}
\bauthor{\bsnm{Lepikhin}, \binits{D.}}, \betal:
\bctitle{{GShard}: Scaling giant models with conditional computation and automatic sharding}.
In: \bbtitle{International Conference on Learning Representations (ICLR)}
(\byear{2021})
\end{bchapter}
\endbibitem

\bibitem[\protect\citeauthoryear{Fedus et~al.}{2022}]{ref19}
\begin{barticle}
\bauthor{\bsnm{Fedus}, \binits{W.}},
\bauthor{\bsnm{Zoph}, \binits{B.}},
\bauthor{\bsnm{Shazeer}, \binits{N.}}:
\batitle{Switch transformers: Scaling to trillion parameter models with simple and efficient sparsity}.
\bjtitle{Journal of Machine Learning Research}
\bvolume{23},
\bfpage{5232}--\blpage{5270}
(\byear{2022})
\end{barticle}
\endbibitem

\bibitem[\protect\citeauthoryear{Dai et~al.}{2021}]{ref20}
\begin{bchapter}
\bauthor{\bsnm{Dai}, \binits{X.}},
\bauthor{\bsnm{Chen}, \binits{Y.}},
\bauthor{\bsnm{Xiao}, \binits{B.}},
\bauthor{\bsnm{Chen}, \binits{D.}},
\bauthor{\bsnm{Liu}, \binits{M.}},
\bauthor{\bsnm{Yuan}, \binits{L.}}:
\bctitle{Dynamic head: Unifying object detection heads with attentions}.
In: \bbtitle{Proceedings of the IEEE/CVF Conference on Computer Vision and Pattern Recognition (CVPR)}
(\byear{2021})
\end{bchapter}
\endbibitem

\bibitem[\protect\citeauthoryear{Cai and Vasconcelos}{2018}]{ref21}
\begin{bchapter}
\bauthor{\bsnm{Cai}, \binits{Z.}},
\bauthor{\bsnm{Vasconcelos}, \binits{N.}}:
\bctitle{Cascade {R-CNN}: High quality object detection and instance segmentation}.
In: \bbtitle{Proceedings of the IEEE Conference on Computer Vision and Pattern Recognition (CVPR)}
(\byear{2018})
\end{bchapter}
\endbibitem

\bibitem[\protect\citeauthoryear{Zhu et~al.}{2022}]{ref22}
\begin{bchapter}
\bauthor{\bsnm{Zhu}, \binits{J.}}, \betal:
\bctitle{{Uni-Perceiver-MoE}: Learning sparse generalist models with conditional {MoEs}}.
In: \bbtitle{Advances in Neural Information Processing Systems (NeurIPS)}
(\byear{2022})
\end{bchapter}
\endbibitem

\bibitem[\protect\citeauthoryear{Eigen et~al.}{2013}]{ref23}
\begin{botherref}
\oauthor{\bsnm{Eigen}, \binits{D.}},
\oauthor{\bsnm{Ranzato}, \binits{M.}},
\oauthor{\bsnm{Sutskever}, \binits{I.}}:
Learning factored representations in a deep mixture of experts.
arXiv preprint arXiv:1312.4314
(2013)
\end{botherref}
\endbibitem

\bibitem[\protect\citeauthoryear{Singh et~al.}{2018}]{ref24}
\begin{bchapter}
\bauthor{\bsnm{Singh}, \binits{B.}},
\bauthor{\bsnm{Najibi}, \binits{M.}},
\bauthor{\bsnm{Davis}, \binits{L.S.}}:
\bctitle{{SNIPER}: Efficient multi-scale training}.
In: \bbtitle{European Conference on Computer Vision (ECCV)}
(\byear{2018})
\end{bchapter}
\endbibitem

\bibitem[\protect\citeauthoryear{Li et~al.}{2019}]{ref25}
\begin{bchapter}
\bauthor{\bsnm{Li}, \binits{Y.}},
\bauthor{\bsnm{Chen}, \binits{Y.}},
\bauthor{\bsnm{Wang}, \binits{N.}},
\bauthor{\bsnm{Zhang}, \binits{Z.}}:
\bctitle{Scale-aware trident networks for object detection}.
In: \bbtitle{Proceedings of the IEEE/CVF International Conference on Computer Vision (ICCV)}
(\byear{2019})
\end{bchapter}
\endbibitem

\bibitem[\protect\citeauthoryear{Zhang et~al.}{2023}]{ref26}
\begin{botherref}
\oauthor{\bsnm{Zhang}, \binits{H.}},
\oauthor{\bsnm{Qiu}, \binits{Y.}},
\oauthor{\bsnm{Wang}, \binits{X.}},
\oauthor{\bsnm{Bai}, \binits{Y.}}:
{UniHead}: Unifying multi-perception for object detection heads.
arXiv preprint arXiv:2309.13242
(2023)
\end{botherref}
\endbibitem

\bibitem[\protect\citeauthoryear{Ultralytics}{2024}]{ref27}
\begin{botherref}
\oauthor{\bsnm{Ultralytics}}:
{YOLO11}: Real-time object detection and image segmentation.
GitHub repository.
[Online]. Available: \url{https://github.com/ultralytics/ultralytics}
(2024)
\end{botherref}
\endbibitem

\bibitem[\protect\citeauthoryear{Zhu et~al.}{2016}]{ref28}
\begin{bchapter}
\bauthor{\bsnm{Zhu}, \binits{Z.}},
\bauthor{\bsnm{Liang}, \binits{D.}},
\bauthor{\bsnm{Zhang}, \binits{S.}},
\bauthor{\bsnm{Huang}, \binits{X.}},
\bauthor{\bsnm{Li}, \binits{B.}},
\bauthor{\bsnm{Hu}, \binits{S.}}:
\bctitle{Traffic-sign detection and classification in the wild}.
In: \bbtitle{Proceedings of the IEEE Conference on Computer Vision and Pattern Recognition (CVPR)},
pp. \bfpage{2110}--\blpage{2118}
(\byear{2016})
\end{bchapter}
\endbibitem

\bibitem[\protect\citeauthoryear{Lin et~al.}{2014}]{ref29}
\begin{bchapter}
\bauthor{\bsnm{Lin}, \binits{T.-Y.}},
\bauthor{\bsnm{Doll{\'a}r}, \binits{P.}},
\bauthor{\bsnm{Girshick}, \binits{R.}},
\bauthor{\bsnm{He}, \binits{K.}},
\bauthor{\bsnm{Hariharan}, \binits{B.}},
\bauthor{\bsnm{Belongie}, \binits{S.}}, \betal:
\bctitle{Microsoft {COCO}: Common objects in context}.
In: \bbtitle{European Conference on Computer Vision (ECCV)},
pp. \bfpage{740}--\blpage{755}
(\byear{2014}).
\bcomment{Springer}
\end{bchapter}
\endbibitem

\end{thebibliography}
\end{document}